# Accelerating material discovery with a threshold-driven hybrid acquisition policy-based Bayesian optimization

Ahmed Shoyeb Raihan[a], Hamed Khosravi[a], Srinjoy Das[b], Imtiaz Ahmed[a]

[a]Department of Industrial and Management Systems Engineering, West Virginia University, Morgantown, WV 26506
[b]School of Mathematical and Data Sciences, West Virginia University, Morgantown, WV 26506

**Abstract**

Advancements in materials play a crucial role in technological progress. However, the process of discovering and developing materials with desired properties is often impeded by substantial experimental costs, extensive resource utilization, and lengthy development periods. To address these challenges, modern approaches often employ machine learning (ML) techniques such as Bayesian Optimization (BO), which streamline the search for optimal materials by iteratively selecting experiments that are most likely to yield beneficial results. However, traditional BO methods, while beneficial, often struggle with balancing the trade-off between exploration and exploitation, leading to sub-optimal performance in accelerated material discovery processes. This paper introduces a novel Threshold-Driven UCB-EI Bayesian Optimization (TDUE-BO) method, which dynamically integrates the strengths of Upper Confidence Bound (UCB) and Expected Improvement (EI) acquisition functions to optimize the material discovery process. Unlike the classical BO, our method focuses on efficiently navigating the high-dimensional material design space (MDS). TDUE-BO begins with an exploration-focused UCB approach, ensuring a comprehensive initial sweep of the MDS. As the model gains confidence, indicated by reduced uncertainty, it transitions to the more exploitative EI method, focusing on promising areas identified earlier. The UCB-to-EI switching policy dictated guided through continuous monitoring of the model uncertainty during each step of sequential sampling results in navigating through the MDS more efficiently while ensuring rapid convergence. The effectiveness of TDUE-BO is demonstrated through its application on three different material science datasets, showing significantly better approximation and optimization performance over traditional EI and UCB-based BO methods in terms of the RMSE scores and convergence efficiency, respectively.

## 1. Introduction

Scientific advancement in the field of material science is entering its fourth paradigm, marked by the integration of big data and artificial intelligence [1]. Initially, the study of materials relied solely on intuitive observation and judgement without scientific quantification [2]. This gradually evolved with the introduction of mathematical models like thermodynamics, providing a theoretical foundation [3]. Following this, the advent of computers enabled simulations of complex problems, with the emergence of methods like density functional theory and molecular dynamics [4]. Despite these significant advancements, the actual process of manufacturing, discovering, and developing new materials with desired properties is still plagued with high experimental costs as well as significant time and resource consumption [5]. There are several reasons behind this slow progress. First and foremost, in most of the problems related to material development and discovery, the underlying relationship between the process parameters or conditions and the desired material properties is very complex [6]. Additionally, during the process of finding the optimal set of these parameters that result in some desired material properties, a high dimensional design space with innumerous combinations must be searched meticulously [7]. Traditional processes require a plethora of physical experiments to be conducted before a desired property can be achieved. However, such processes often cannot generate desired solutions as these physical experiments are expensive and requires significant effort and time [8], [9]. Moreover, the nature of current research methods for manufacturing and developing material with desired properties stems from the heavy reliance on human involvement, which further impedes the process of accelerated discovery.

The current shift to the fourth paradigm, driven by the advent and widespread use of artificial intelligence aims to resolve these issues through data-driven discovery processes. The intersection of materials science and computer science, particularly with the application of machine learning, is transforming the field, allowing for faster and more efficient predictions of material properties and the discovery of new materials [10], [11]. This transformation is fueled by the vast amount of data generated from experiments and computational methods, accelerated by high-throughput computing [12]. Moreover, in light of growing sustainability concerns, including reducing carbon footprints, minimizing resource and energy use, waste reduction, and preserving critical materials, scientists and researchers are increasingly focused on data-driven approaches to accelerate the process of material discovery [13]. The material discovery process is essentially an optimization challenge where the aim is to enhance or reduce certain properties of a material [14]. This is achieved by adjusting specific features or parameters, which are influenced by the material's chemistry and processing conditions. As we dive deeper, we observe that the process of identifying the optimal set of these parameters which results in some desired material property is equivalent to solving the global optimization problem, often represented mathematically as [15]:

$$x^* = arg \max_{x \in \chi} f(x) \tag{1}$$

where $x^*$ represents the specific action or optimal set parameters that optimizes the function $f(x)$. This is like searching for a needle in a haystack, given the high dimensional material design space (MDS) denoted by $\chi$ with countless combinations or inputs denoted by $x$. The challenge here is that the function $f(x)$ – indicative of the outcome of an action or process parameter – lacks an explicit form and is termed as black-box function [16]. To decipher $f(x)$, material scientists and engineers have traditionally resorted to physical experiments at certain MDS locations. However, these experiments, pivotal for understanding the landscape of material properties, are both resource-intensive and time-consuming [17]. A transformative approach to materials discovery and development is thus necessary, one that requires intelligently navigating through the vast MDS and finding configurations that align with the desired target properties. Such an approach, presumably autonomous in nature, involves navigating the MDS by adeptly balancing between exploitation, which means taking opportunistic steps for immediate improvement based on current knowledge, and exploration, which pertains to venturing into unexplored regions of the MDS to possibly gain better rewards. Under resource constrained environments, where conducting multiple physical experiments are deemed infeasible, achieving the proper balance between exploitation and exploration may be the only way to discover and develop materials with desired properties efficiently. Consequently, active or sequential learning has emerged offering a structured method to determine the most effective combination of process parameters in order to iteratively select and obtain the desired material property efficiently and effectively [18]. Perhaps, the most popular active learning strategy is Bayesian Optimization (BO), acclaimed for its potential to revolutionize material discovery, development, and optimization [19]. By fusing data-driven methodologies with domain expertise, BO-based active learning techniques endeavor to efficiently explore and pinpoint optimal materials through a synergetic relationship between a surrogate model and an acquisition function [20], [21]. As a result, BO and its variants have found widespread application in a variety of research areas including material discovery and development [5], [13], [15], [22], drug discovery and pharmacology [23]–[25], robotics [26], [27], aerospace engineering [28], [29], automotive industry [30], [31], chemical science [32]–[34] and so on.

In the BO literature, while Gaussian Process (GP) has always been the most obvious choice as the surrogate model for $f(x)$, there has been much debate about the choice of acquisition functions which enable the determination of $x^*$ as per Equation 1 [35]. The performance of these sequential learning algorithms almost always boils down to achieving the balance between the exploitation and exploration of the MDS. Too much exploitation causes the algorithm to get trapped in local optima whereas too much exploration can cause the algorithm to waste valuable resources in less promising regions of the MDS without finding the optimum value or desired property [36]. Two of the most popular acquisition functions in the BO literature are Expected Improvement (EI) and the Upper Confidence Bound (UCB). While both functions serve to achieve a good exploitation-exploration tradeoff balance, the EI is inherently more exploitative in nature while the UCB is more explorative [37], [38]. In material discovery, where it is crucial to get an overall understanding of the vast MDS besides reaching the optimal value under the constraint of limited resources and time, the EI-based BO and the UCB-based BO by themselves often deliver sub-optimal performance. Keeping the idea about the strengths and weaknesses of both these acquisition functions, in this work we propose to develop a hybrid approach where both UCB and EI are utilized to efficiently explore and exploit the MDS as required. Our proposed approach named as Threshold-Driven UCB-EI BO or in short TDUE-BO aims to guide the sequential experiments by maintaining a better balance between exploration and exploitation. By starting with UCB, TDUE-BO ensures adequate exploration of the design space, which is particularly important in the initial phases of optimization to avoid missing out on unexplored high-potential regions. Once the uncertainty falls below a certain threshold, indicating that the model has a reliable understanding of the function's behavior, the strategy switches to EI. This shift allows for a more focused exploitation of the most promising regions identified during the exploration phase. We have applied our proposed methodology in three different material science datasets and achieved significantly better performance in terms of RMSE scores compared to when we applied EI-based BO and UCB-based BO separately in these datasets. We have also shown how the threshold-guided switching between the UCB and EI in the sequential learning environment can generate better approximation of the MDS by providing the iterative learning of a one-dimensional function with TDUE-BO versus the other two conventional approaches.

The rest of the paper is organized as follows. Section 2 provides a short discussion on the three material datasets we have used in our work. The working of the traditional BO with GP as the surrogate model, and EI and UCB as acquisition functions is presented in Section 3 with our proposed active learning framework (TDUE-BO). In Section 4, we first present an illustrative example showing how TDUE-BO adaptively switches between UCB and EI to optimize and approximate an unknown function outperforming both the EI and UCB-based BO. Next, we present the results of our proposed approach, along with the competing EI and UCB guided BO approaches when applied to the three material datasets. We analyze our findings at the end of this section. Finally, we conclude our work in Section 5, providing directions for future research.

## 2. Data Understanding

In this study, we considered three datasets from three distinct material domains, each with its own size, dimension, and material system characteristics [18]. There are multiple materials science domains represented in these datasets, each with an optimization goal and specific synthesis method. The first dataset is the P3HT/CNT dataset where the target property is the 'Electrical Conductivity' has 233 samples and was synthesized using the drop-casting method. Residing in the domain of composite blends, this dataset has 5 input features [39]. The second dataset is the Perovskite dataset with 3 input features (process parameters) and the output target is the 'Instability Score' [40]. Belonging to the domain of thin-film perovskite, this dataset contains 139 data points for these input features and their corresponding output. The AutoAM dataset with 100 samples from the field of materials manufacturing is the third dataset we have used in our work [41] and it is related to 3D printing. This dataset has 4 process parameters, and the 'Shape Score' is the output target. All the three datasets are derived from high-throughput experimental platforms.

As a result of the diversity of these datasets, we can assess the effectiveness of our proposed method within the field of experimental materials science, emphasizing its versatility and adaptability to a wide range of high-throughput experiments. In order to improve our analysis and gain a comprehensive perspective, we conducted a detailed exploration of the data. To capture the predominant variance observed across the datasets, we used the largely used Principal Component Analysis (PCA) methodology to analyse the complexity of these datasets into two principal components, PC 1 and PC 2 [42]. Through this reduction, we are able to visualize and interpret the datasets more easily, enabling us to identify the patterns and relationships pertinent to our study, which focuses on the optimization objectives.

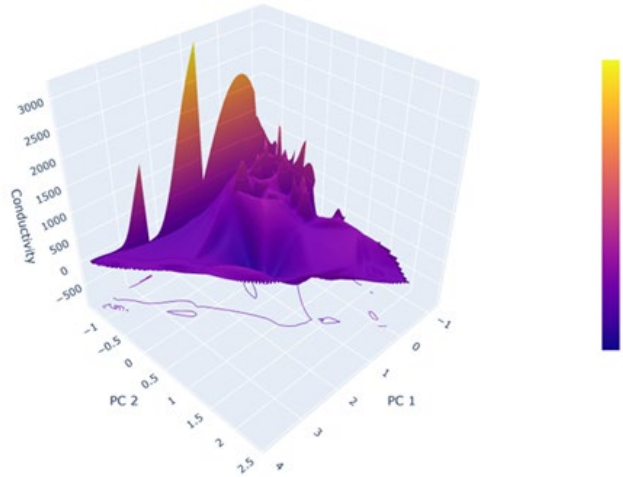

Figure 1: 3D surface representation of 'Conductivity' distribution across the principal components in the P3HT/CNT dataset

Using PCA, Figure 1 illustrates the transformed feature space of the P3HT/CNT dataset, focusing on the 5 features 'P3HT content (%)', 'D1 content (%)', 'D2 content (%)', 'D6 content (%)', and 'D8 content (%)'. Upon examination of the plot, there appears to be a concentration of data points around the highest peak, which possibly represents the optimal material properties. As conductivity (target property) values increase, the color gradient transitions from cool to warm colors, indicating an increase in conductivity. In particular, the warmest color zones are aligned with the peaks of the plot, suggesting that these regions correspond to higher conductivity levels. The same analysis has been conducted for the two other datasets, as well. Figure 2 shows the feature space for the Perovskite dataset where the dynamic range of values for the target 'Instability Index' shows the variable nature of the materials used in the process. With its multiple peaks and troughs, this surface reaches a maximum stability index of about 1.2M at the highest peak, which contrasts strikingly with lower index values of nearly 0.8M at the lower peaks. This color spectrum reflects the degree of instability, shifting from a deep, rich purple at the lower end of the spectrum to a bright, intense orange at the upper end. The warmer colors indicate higher levels of instability. Throughout the base are contour lines showing a distribution of data points at varying levels of 'Instability Index', indicating a non-uniform distribution of material properties. There may be a concentration of points in an area where the contour lines are densely packed, which may indicate a common manufacturing setting or material composition, while sparser contour lines may indicate less explored or more variable conditions.

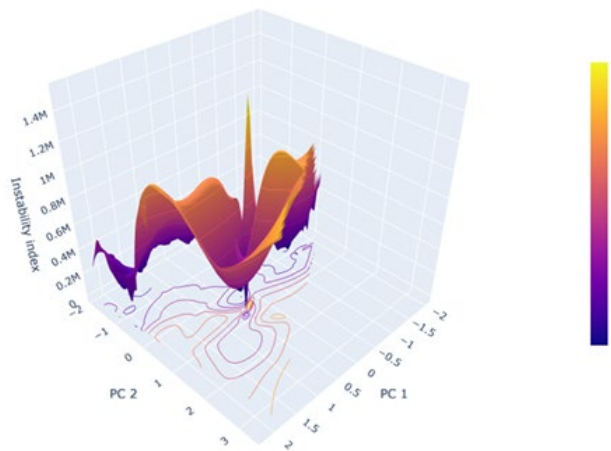

Figure 2: 3D surface representation of 'Instability Index' distribution across the principal components in the Perovskite dataset

Finally, Figure 3 shows the 3D surface plot for the AutoAM dataset which indicates the parameters that have been evaluated, including the 'Prime Delay', 'Print Speed', 'X Offset Correction', and 'Y Offset Correction' with the aim of optimizing the 'Score'. As can be seen in the surface plot, there is a prominent peak in the center of the plot, which quickly increases to a score of approximately 60, thereby indicating a concentration of data points around which optimal manufacturing conditions exist. The peak is surrounded by a plateau of moderate scores that range from 20 to 40, beyond which the scores gradually decrease to near zero levels as one moves further out along the PC 1 and PC 2 axes. 'Score' spectrum, where purple is located at the bottom and yellow is located at the top, has a vibrant gradient transition that visually emphasizes the regions of high and low scores. In addition to illustrating the density and spread of data points, the contour lines extending across the base illustrate the variability within the manufacturing process parameters. By using this graphical representation, we are not only able to simplify the multivariate analysis but also to identify areas where conductivity is enhanced, thereby providing a strategic direction for future experiments and material development.

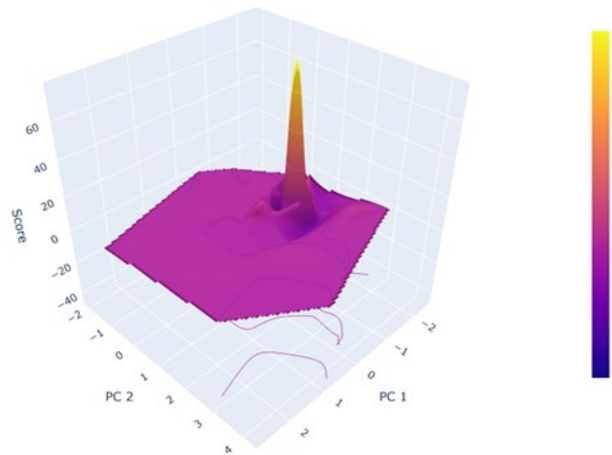

Figure 3: *3D surface representation of 'Score' distribution across the principal components in the AutoAM dataset*

## 3. Methodology

BO is an efficient method for the optimization of black-box functions that are expensive or time-consuming to evaluate [17]. It is particularly useful in scenarios where the objective function does not have a closed-form expression, is costly to evaluate, and where derivatives are not available or are unreliable. The core concept behind BO is to use a probabilistic model to estimate the function. This model is used to make predictions about the function's behavior and to quantify the uncertainty in these predictions. The probabilistic nature of the model makes it well-suited to guiding the search for the optimum in a principled and efficient manner. The process of BO is iterative; it alternates between modeling and optimizing which calls for dividing its framework into two parts: a surrogate function for modeling and an acquisition function for optimizing [20]. During each iteration, BO attempts to approximate the unknown function $f$ by refining the surrogate model and uses the acquisition function to determine the next design point to assess.

*3.1. Surrogate Model*

A surrogate model is a predictive model that approximates the objective function of interest. The key purpose of a surrogate model is to provide a computationally cheap and analytically tractable estimate of the objective function, which is typically expensive, time-consuming, or difficult to evaluate directly. Surrogate models approximate the behavior of the complex objective function based on available data (sampled points). They predict what the objective function's output would be at unsampled points. Besides providing predictions, good surrogate models also quantify the uncertainty or confidence in their predictions. This uncertainty plays a crucial role in decision-making in BO. Gaussian Processes (GPs) are a popular choice for surrogate models in BO due to their robustness and effectiveness in capturing the uncertainty of predictions. A Gaussian Process is a collection of random variables, any finite number of which have a joint Gaussian distribution [43]. It is fully specified by its mean function $m(x)$ and covariance function $k(x, x')$ which is given by the following equation:

$$f(x) \sim GP(m(x), k(x, x')) \tag{2}$$

The mean function represents the average value of the function, and the covariance function, also known as the kernel, defines the similarity between different points in the input space. For a set of input points $X_*$ where predictions are required, and given a set of training points $X$ with corresponding target values $y$, the joint distribution of the observed targets $y$ and the function values at the new input points $f_*$ under the GP prior is:

$$\begin{bmatrix} y \\ f_* \end{bmatrix} \sim N\left(\begin{bmatrix} m(X) \\ m(X_*) \end{bmatrix}, \begin{bmatrix} K(X, X) & K(X, X_*) \\ K(X_*, X) & K(X_*, X_*) \end{bmatrix}\right) \tag{3}$$

The predictive distribution for $f_*$ is then a Gaussian with mean $\mu_*$ and covariance $\Sigma_*$, which is given by:

$$\mu_* = m(X_*) + K(X_*, X)[K(X, X) + \sigma_n^2 I]^{-1}(y - m(X)) \tag{4}$$

$$\Sigma_* = K(X_*, X_*) - K(X_*, X)[K(X, X) + \sigma_n^2 I]^{-1} K(X, X_*) \tag{5}$$

where, $\sigma_n^2$ is the noise term added to the diagonal of $K(X, X)$ for numerical stability and to account for observation noise. One of the key advantages of GPs is their ability to provide not only predictions but also quantify uncertainty in these predictions, a feature crucial for effective exploration-exploitation trade-offs in optimization. As non-parametric models, GPs adaptively increase their complexity with the availability of more data, making them suitable for a wide range of functions, from simple to highly complex. The use of kernel functions in GPs allows for the incorporation of prior knowledge and assumptions about the function's behavior, such as smoothness or periodicity, enhancing their adaptability [44]. Additionally, GPs offer analytical tractability, enabling closed-form expressions for predictions and uncertainties, which are computationally efficient.

*3.2. Acquisition Function*

In BO, the acquisition function plays a pivotal role in guiding the sampling process. It is essentially a strategy or heuristic used to decide where to sample next, balancing the need for exploration (sampling in areas with high uncertainty) and exploitation (sampling in areas where the model predicts accurate values) [16]. Using the uncertainty information provided by the surrogate model, the acquisition function helps to make informed decisions about where to sample next. There are many popular choices for acquisition functions in the literature of BO. However, Expected Improvement (EI) and Upper Confidence Bound (UCB) are two popularly used acquisition functions [45].

*Expected Improvement (EI):* EI is a widely used acquisition function in Bayesian Optimization (BO) for guiding the selection of the next sampling point. It is particularly effective in striking a balance between exploring new areas and exploiting the known promising regions of the search space. EI focuses on regions of the input space where there's the highest expected improvement over the current best observation [16]. It essentially quantifies the expected amount by which a given point $x$ could improve over the current best-known value $f(x^+)$ [37]. EI at a point $x$ is defined as the expectation of the improvement function $I(x)$, which is the positive difference between the current best value $f(x^+)$ and the potential function value at $x$. Mathematically, it's expressed as:

$$I(x) = \max(0, f(x^+) - f(x)) \tag{6}$$

The EI at point $x$ under a Gaussian Process (GP) model is then given by:

$$EI(x) = E[I(x)] \qquad (7)$$

For a GP, this expectation can be calculated analytically. Assuming the GP prediction at point $x$ is normally distributed with mean $\mu(x)$ and standard deviation $\sigma(x)$, the EI can be computed as:

$$EI(x) = (\mu(x) - f(x^+) - \xi)\Phi(Z) + \sigma(x)\phi(Z) \qquad (8)$$

where, $\mu(x)$ and $\sigma(x)$ are the mean and standard deviation of the GP's predictions at $x$, $\Phi$ and $\phi$ are the cdf and pdf of the standard normal distribution, respectively, $\xi$ is a small non-negative parameter that introduces a trade-off between exploitation and exploration. $Z$ is a standardization of the improvement function given by:

$$Z = \frac{\mu(x) - f(x^+) - \xi}{\sigma(x)} \qquad (9)$$

*Upper Confidence Bound (UCB):* UCB is another key acquisition function used in Bayesian Optimization (BO), particularly known for its ability to balance exploration and exploitation efficiently. Unlike some other acquisition functions that primarily focus on regions of high expected value or improvement, UCB explicitly incorporates the uncertainty of the surrogate model predictions into the decision-making process. UCB is based on the principle of "optimism in the face of uncertainty" [46]. It selects the next point to sample by considering not only the expected value of the function at each point (exploitation) but also the uncertainty or variance associated with that prediction (exploration) [47]. The idea is to sample points where the surrogate model predicts high values or where the uncertainty of the prediction is high, thus potentially discovering better maxima than currently known. Mathematically, the UCB at a point $x$ in the input space is defined as a sum of the predicted mean and a confidence interval around the mean:

$$UCB(x) = \mu(x) + \kappa \cdot \sigma(x) \qquad (10)$$

where, $\mu(x)$ and $\sigma(x)$ are the mean prediction and the standard deviation or the uncertainty of the surrogate model, respectively at point $x$, and $\kappa$ is a tunable parameter that determines the trade-off between exploration and exploitation.

*3.3. Proposed Methodology (TDUEBO)*

EI tends to focus on regions close to the best observed outputs, making it inherently exploitative. While this is beneficial for refining the search around promising areas, it can lead to premature convergence and missing out on potentially better regions that haven't been sufficiently explored. On the other hand, UCB, with its emphasis on uncertainty, inherently leans towards exploration. While this is advantageous for discovering new regions of the search space, it can result in excessive exploration, especially in later stages of optimization when refining the solution is more critical. In our proposed framework, these limitations are addressed by intelligently combining UCB and EI in a hybrid acquisition function strategy which switches from being more explorative in the initial phase (UCB-based) to being more exploitative (EI-based) in the later phase. We employ a threshold guided strategy to dictate the dynamic UCB-EI switching during the closed-loop active learning process. The rationale for using a threshold-driven approach in BO hinges on effectively balancing the two fundamental aspects of the optimization process: exploring the design space to find regions of potential interest and exploiting known regions to refine the search around promising areas. Initially, when the surrogate model has limited data, the uncertainty across the design space is high. Using UCB initially emphasizes exploration, which is vital to avoid local optima and discover diverse regions of potential interest. As more data points are gathered, the model's uncertainty decreases, and a more exploitation-focused strategy becomes beneficial. The threshold for switching from UCB to EI is based on the level of average uncertainty in the model's predictions. When the average uncertainty drops below a certain level, it indicates that the model has a relatively more confident understanding of the function's behavior. At this point, switching to EI is rational because it focuses more on exploiting the regions around the current best observations, which is more efficient once the model is sufficiently informed. We resort to domain knowledge and expertise to guide the setting of this threshold. Based on the property of the process parameters and the desired output, the threshold is set to reflect the point at which further exploration is unlikely to yield practically significant improvements. The average uncertainty or standard deviation across the input space in a GP model can be calculated by averaging the standard deviation of the GP's predictions at a set of points. This average gives an overall measure of the model's uncertainty about the function it's approximating. For any point $x_*$, the variance can be calculated using the GP model as Equation 11. The standard deviation or uncertainty for $x_*$ is calculated simply taking the square root of this variance which is given by Equation 12.

$$\sigma^2(x_*) = K(x_*, x_*) - K(x_*, X)[K(X, X) + \sigma_n^2 I]^{-1} K(X, x_*) \tag{11}$$

$$\sigma(x_*) = \sqrt{\sigma^2(x_*)} \tag{12}$$

Finally, the average uncertainty or average standard deviation across all the $N$ points ($X_*$) is calculated using Equation (13) where $N$ is the number of points in $X_*$, and $x_{*i}$ represents the $i$-th point in $X_*$. We have divided our proposed methodology into five unique steps which are described in the following:

$$\bar{\sigma} = \frac{1}{N} \sum_{i=1}^{N} \sigma(x_{*i}) \tag{13}$$

**Step 1 (Initialization)**: Start with a set of initial samples, $D_i = \{X_i, y_i\}$ to fit the GP model using Equation (2). The initial samples are obtained after dividing each dataset into three parts. One part is used to fit the initial GP model, another part is used as the candidate locations for the sequential updates during the closed-loop BO process. The third part is reserved to test the performance of all the approaches. For instance, for the P3HT/CNT dataset, out of 233 original samples, 30 were used for initial model fitting, 144 samples were allocated for the sequential experiment candidates and the remaining 59 samples were kept for testing the approaches.

**Step 2 (Sequential Sampling with UCB)**: Begin the active learning or sequential experiment process using UCB as the acquisition function using. This phase emphasizes exploring the design space to uncover regions with high uncertainty or potential. The available candidate samples for this sequential process are selected from the second part of the original dataset which can be represented by $D_{seq} = \{X_{seq}, y_{seq}\}$. In the P3HT/CNT dataset, out of 144 allocated candidate samples, 50 samples were used for updating the models sequentially. This means that considering the initial 30 samples and now 50 sequential samples, the experimental budget for the P3HT/CNT dataset is limited to 80 samples.

**Step 3 (Monitoring Uncertainty)**: Continuously monitor the uncertainty in the GP model's predictions. This involves assessing the average standard deviation of the predictions across the design space using Equations (11-13).

**Step 4 (Switching to EI)**: Once the uncertainty falls below a predefined threshold, switch the acquisition function from UCB to EI. This marks the transition to a more exploitative phase, focusing on refining the search in the most promising areas.

**Step 5 (Iterative Update)**: Continue the iterative process of selecting new sample points based on EI from the set $D_{seq}$, updating the GP model, and re-evaluating until our experimental budget is exhausted. In this work, this experimental budget varies depending on the dataset we use. The framework of TDUE-BO is illustrated in Figure 4. Figure 5 shows the pseudocode of the proposed TDUE-BO algorithm.

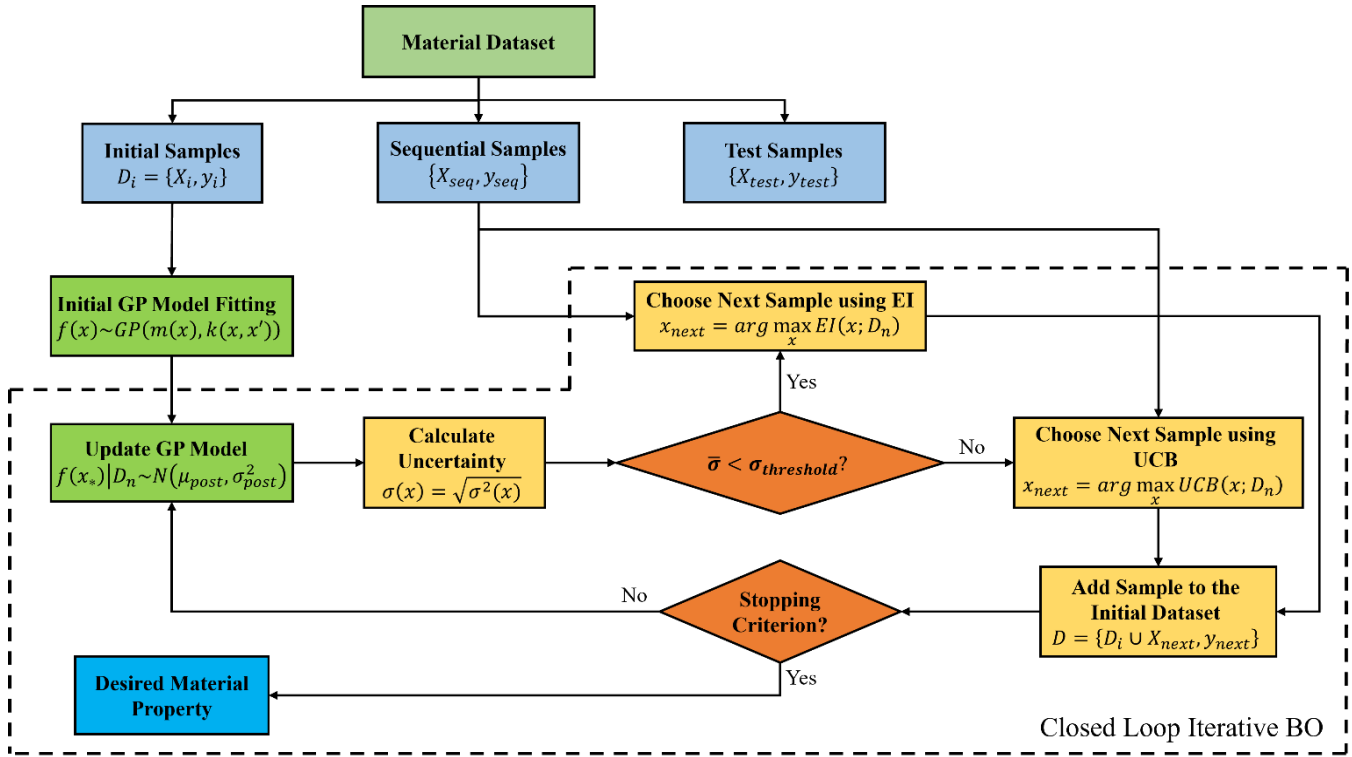

Figure 4: Flow diagram of the proposed TDUE-BO framework

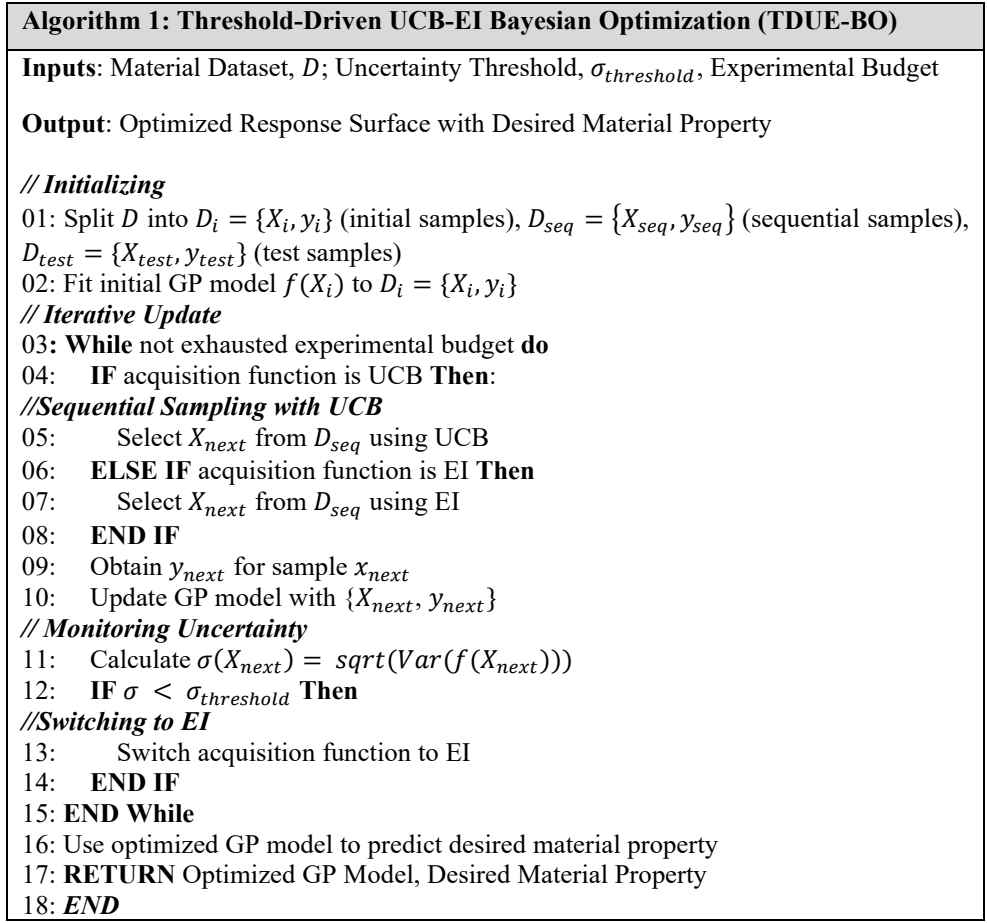

Figure 5: Pseudocode of the proposed TDUE-BO sequential learning approach

## 4. Results and Discussion

*4.1. An Illustrative Example*

To visualize the working of our proposed framework, in this section, we have attempted to approximate a simple univariate function using our approach. We have also shown how our methodology performs better in approximating an unknown function than the widely used EI-based and UCB-based BO frameworks. The univariate function that we attempted to predict has a very simple form which can be expressed by Equation (14) where $f(\mathbf{x}) = -\sin 3x - x^2 + 0.7x$ and $\xi$ is used to denote a Gaussian noise.

$$y = f(\mathbf{x}) + \xi \qquad (14)$$

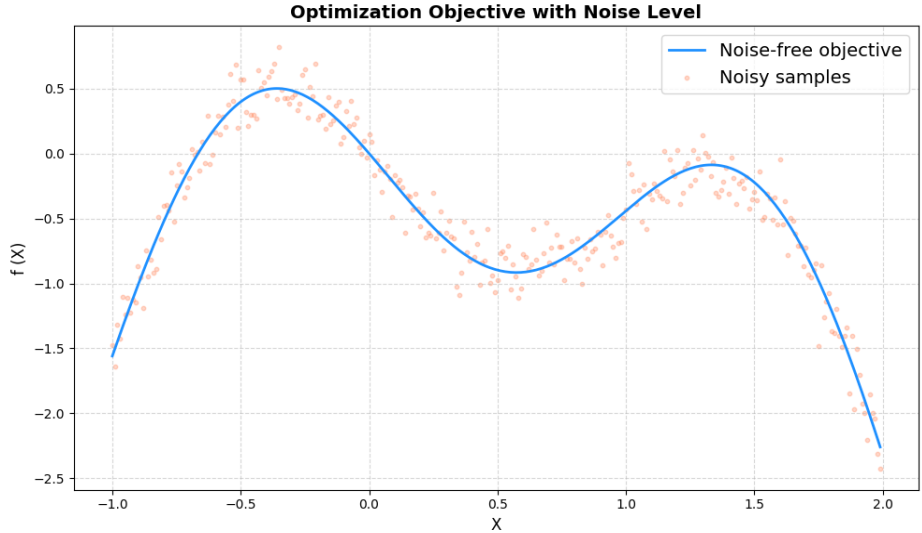

Figure 6: A 1-dimensional function with noisy samples

The function has a very simple structure with two peaks as shown in Figure 2 with the green curve. The noisy samples are the blue dots where the mean and standard deviation of the Gaussian noise is chosen to be 0 and 0.2, respectively. In the absence of knowledge about the underlying true function, our proposed sequential algorithm would start with two initial experiments (one at -0.9 and the other at 1.1) as seen in the figure. This is necessary to construct a statistical model. In simple words, to approximate the above function, we need to start with some known input values along the $x$ axis and their corresponding $y$ or $f(x)$ values. The range of the input values in this case is [-1.0, 2.0]. For the sequential updates, we decided to work with an experiment budget of 11 candidate locations or experiments. As the statistical model, we fit a Gaussian Process with Matern kernel using these two initial experiments. We present the approximation of this univariate function with EI-based BO, UCB-based BO and then our proposed hybrid UCB-EI-based BO. Note that we are showing the first and the last 4 iterations of each of this approach. As seen from Figure 7 where we approximated the one-dimensional function, the final approximation using EI-based BO after 12 iterations although identified the maxima of the function, could not however approximate the right most peak of the function (see Iteration 12). We can also see from Figure 7 that EI is more exploitative in nature, overexploiting the region around the left peak of the function and as a result could not give us an overall idea about the entire response surface. In Figure 8, where we approximated the similar function with UCB-based BO, we see that UCB does a better job in giving us a decent understanding of the response surface by exploring more. However, it could not correctly determine the optima. Our proposed TDUE-BO, on the other hand, successfully achieved a balance by doing the right amount of both exploration and exploitation and consequently, we see in the Figure 9 that after 12 iterations, the entire response surface was approximated as well as the optima has been identified.

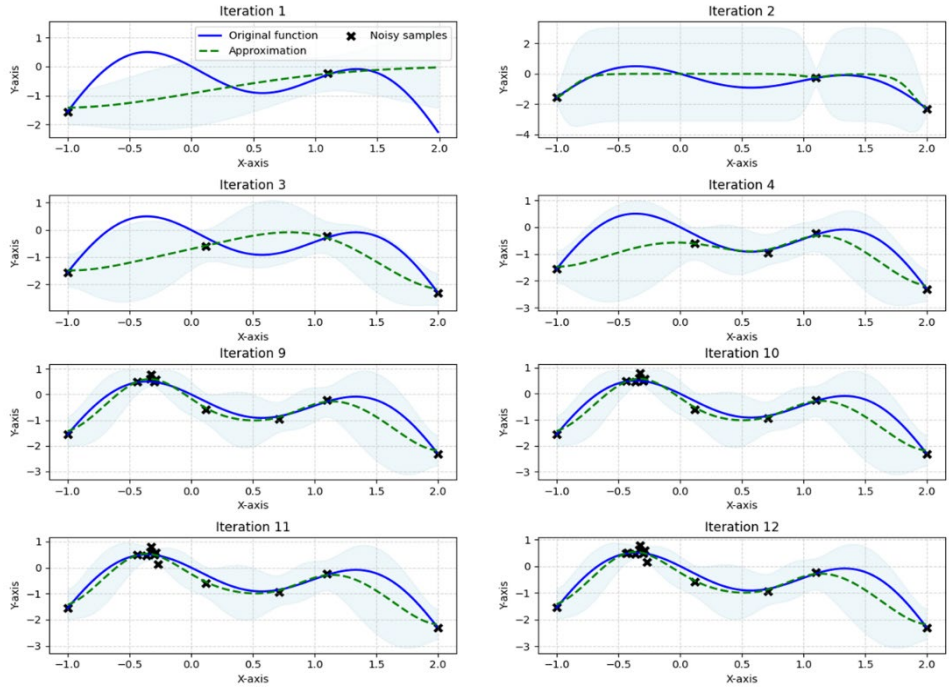

Figure 7: Approximation of a 1-dimensional function with EI-based BO

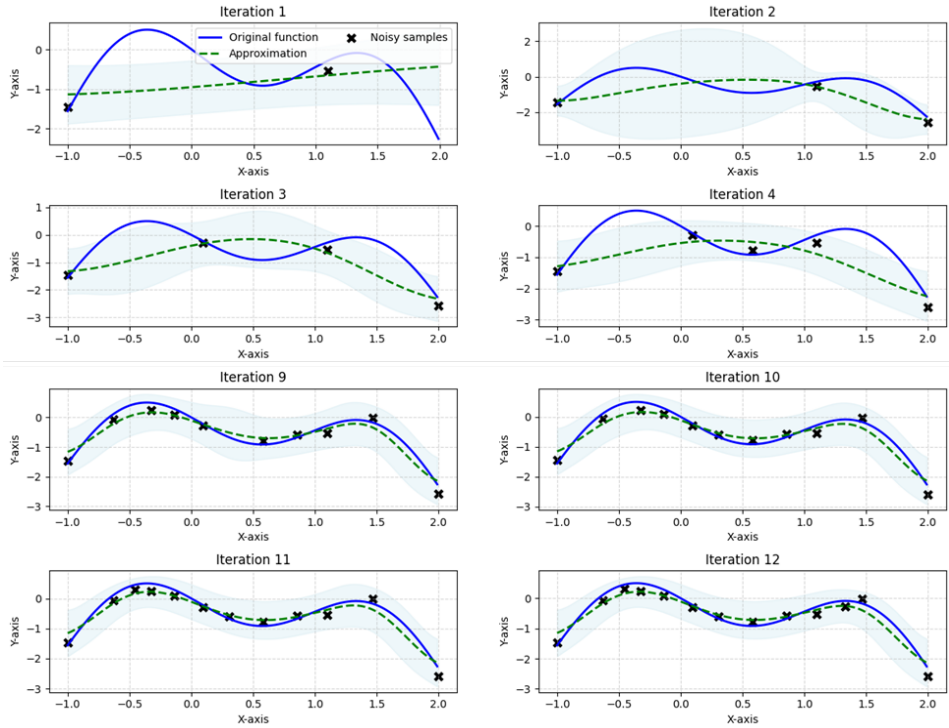

Figure 8: Approximation of a 1-dimensional function with UCB-based BO

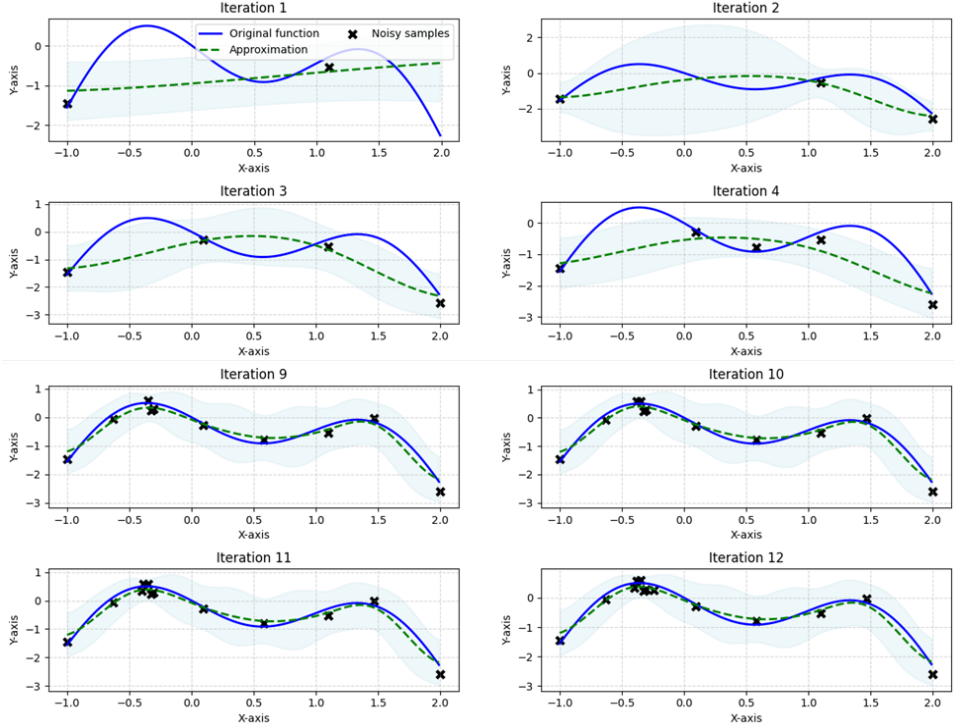

Figure 9: Approximation of a 1-dimensional function with TDUE-BO

*4.2. Application in Materials Datasets*

Our proposed framework, Threshold-Driven-UCB-EI BO (TDUE-BO) is applied in three different material science datasets. All these three datasets (Perovskite, P3HT/CNT, and AutoAM) are obtained through high throughput experiments (HTE). After data normalization has been applied, the datasets became ready for use in our proposed methodology. Table 1 shows the classification of these datasets into training and test sets as well as the number of samples or points used from the training sets to fit the initial GP and to perform the sequential experiments. For instance, the Perovskite dataset has a total of 139 samples. We split it into training (75%) and test (25%) sets. Out of the 104 samples in the training set, we used 20 samples for the initial model fitting using GP and used the remaining 84 samples as candidates for the sequential experiments. Out of these 84 samples, we decided to keep the experimental budget to 40 sequential iterations which indicates that the initial GP model has been updated sequentially 40 times, each time with one sample added to the surrogate model. We followed a similar procedure for the other two datasets (P3HT/CNT and AutoAM). Based on the dataset size, different number of samples were used for the initial model fitting and the sequential iterations.

Table 1: Description of the three datasets

| Dataset | Training Set | Initial Model Fitting | Sequential Experiments | Test Set |
| --- | --- | --- | --- | --- |
| Perovskite | 104 | 20 | 40 (84) | 39 |
| P3HT/CNT | 174 | 30 | 50 (144) | 59 |
| AutoAM | 75 | 15 | 30 (60) | 25 |

We compared the performance of our proposed TDUE-BO framework with the existing UCB-based BO and EI-based BO, respectively. As the performance metric, we employed the root mean square error (RMSE) score defined as the square root of the average of the squared differences between the predicted values and the actual values. It can be expressed as mathematically as:

$$RMSE = \sqrt{\frac{\sum_{i=1}^{n}(y_i - \hat{y}_i)^2}{n}} \tag{15}$$

where, $y_i$ and $\hat{y}_i$ represents the actual observed and the predicted values, respectively and $n$ represents the total number of observations. To reduce the uncertainty while assessing the performance, we have calculated the RMSE scores for these three datasets 30 times. Table 2 compares the mean, median and the interquartile range (IQR) for these three datasets with the EI-BO, UCB-BO and our proposed methodology (TDUE-BO) over 30 RMSE score calculations. A lower RMSE score indicates a better approximation performance. From the table, we can see that for the Perovskite dataset, our method (TDUE-BO) achieved better results in terms of all the metrics (highlighted in red). IQR, measuring the statistical dispersion, is significantly lower (0.015459) which indicates a better consistency and reliability in prediction performance for our approach. For the P3HT/CNT dataset, TDUE-BO outperformed both the EI and UCB-based BO in terms of the mean and median achieving lower RMSE values. Finally, in the AutoAM dataset, TDUE-BO once again achieved superior performance achieving lower values of mean, median and IQR for the RMSE scores.

Table 2: Performance comparison of the three competing approaches for the three datasets using RMSE scores

| Statistic | EI-BO | UCB-BO | TDUE-BO |
|---|---|---|---|
| **Perovskite Dataset** | | | |
| Mean | 0.160957 | 0.137608 | **0.108539** |
| Median | 0.171725 | 0.113114 | **0.109779** |
| IQR | 0.091104 | 0.082315 | **0.015459** |
| **P3HT/CNT Dataset** | | | |
| Mean | 0.12517 | 0.124795 | **0.120858** |
| Median | 0.124201 | 0.12333 | **0.120856** |
| IQR | 0.025107 | **0.024028** | 0.025467 |
| **AutoAM Dataset** | | | |
| Mean | 0.156872 | 0.142149 | **0.134824** |
| Median | 0.15181 | 0.136413 | **0.129976** |
| IQR | 0.058507 | 0.044231 | **0.026041** |

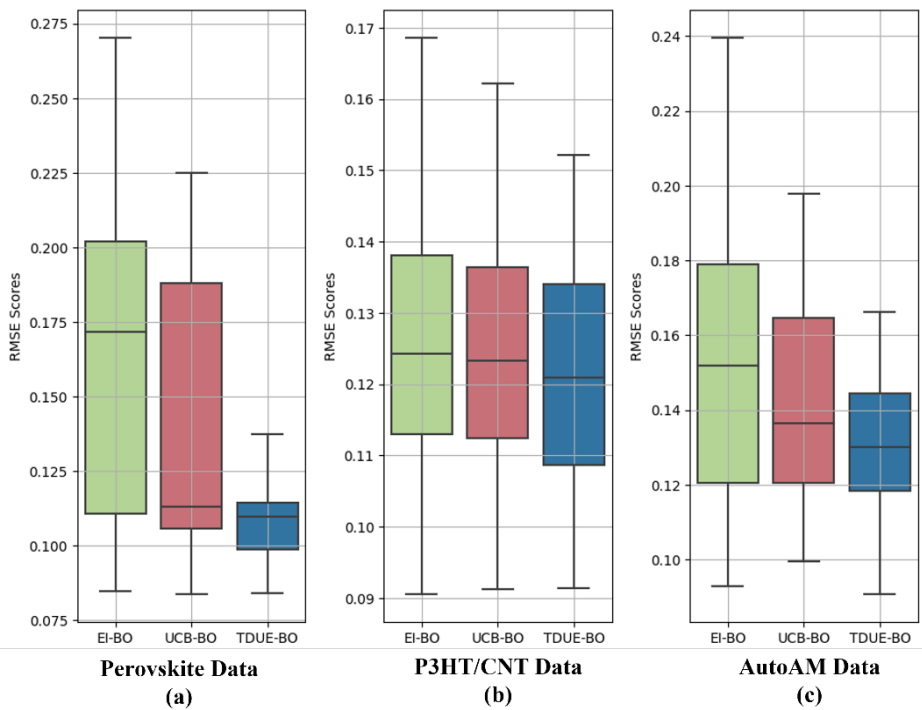

Figure 10: (a) Boxplots for the RMSE scores of three competing approaches for (a) Perovskite dataset (b) P3HT/CNT dataset and (c) AutoAM dataset

The boxplots of the RMSE scores for the three competing approaches are also illustrated in the following figures. Figure 10 (a) shows the boxplot for the Perovskite dataset. We can see that in addition to obtaining a significantly reduced RMSE value, the dispersion of the RMSE values for 30 runs is also much less. From Figure 10 (b), for the P3HT/CNT dataset, the boxplot shows lower RMSE scores for our approach. The boxplot for the AutoAM dataset once again underscores the superior performance of TDUE-BO over the other two approaches in Figure 10 (c).

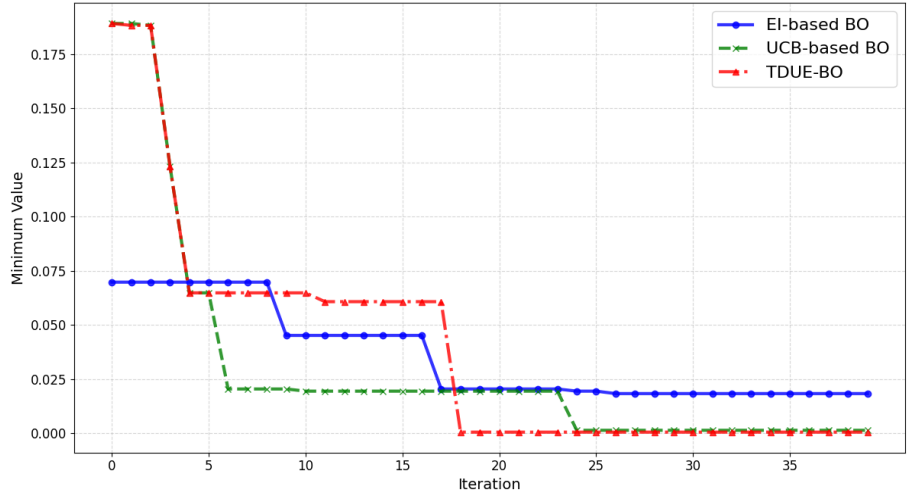

Figure 11: Iterations to reach the minimum for the three competing approaches for the Perovskite dataset

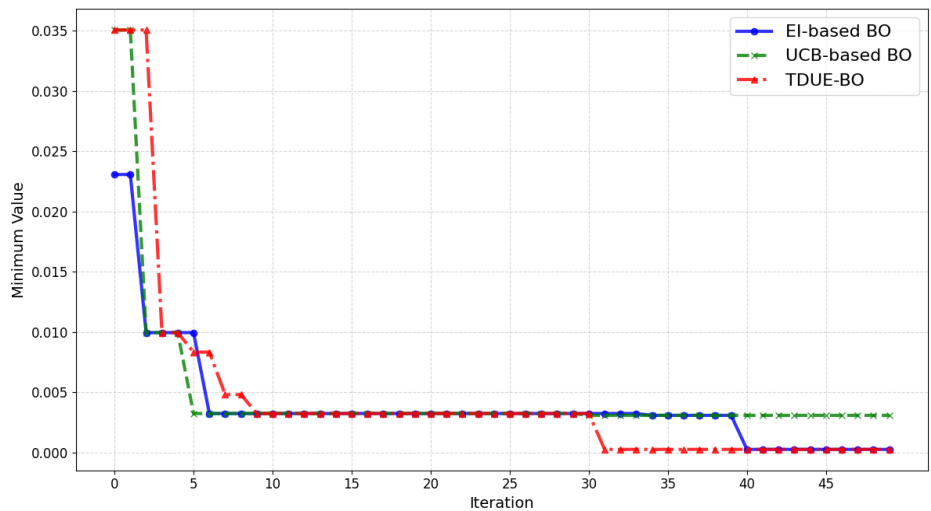

Figure 12: Iterations to reach the minimum for the three competing approaches for the P3HT/CNT dataset

In terms of RMSE score, our proposed approach TDUE-BO does a better job at approximating the underlying relationship between the process parameters and the desired property in the three material datasets. However, convergence is also another crucial aspect in these sequential learning algorithms. Keeping this in mind, we also attempted to analyze the convergence of our approach compared to the two traditional approaches. To achieve this, for each of the datasets, we checked how TDUE-BO performs in determining the minimum value of the target property which is the desired property for all these datasets. The results are shown in the following figures. Figure 11 shows the convergence of all the three methods for the Perovskite dataset. We can see TDUE-BO is able to determine the minimum value within 18 iterations which is the quickest. EI-based BO, in this case, failed to converge even after 40 sequential updates. Continuing from iteration number 17 till 40, EI was exploiting around the same region due to its over-exploiting tendency and therefore could not locate the optimal target property. Figure 12 shows the convergence for the P3HT/CNT dataset. We can once again see the superior convergence of TDUE-BO over the two other methods which successfully reached the minimum within 31 iterations. In this case, UCB-BO, due to prioritizing exploration more than exploitation, failed to identify the minimum value whereas it took 39 iterations for EI-BO to converge. Finally, for the AutoAM dataset, TDUE-BO converged much earlier taking only 13 iterations compared to its counterpart EI-based BO which took 21 iterations. UCB-based BO once again could not converge for this dataset.

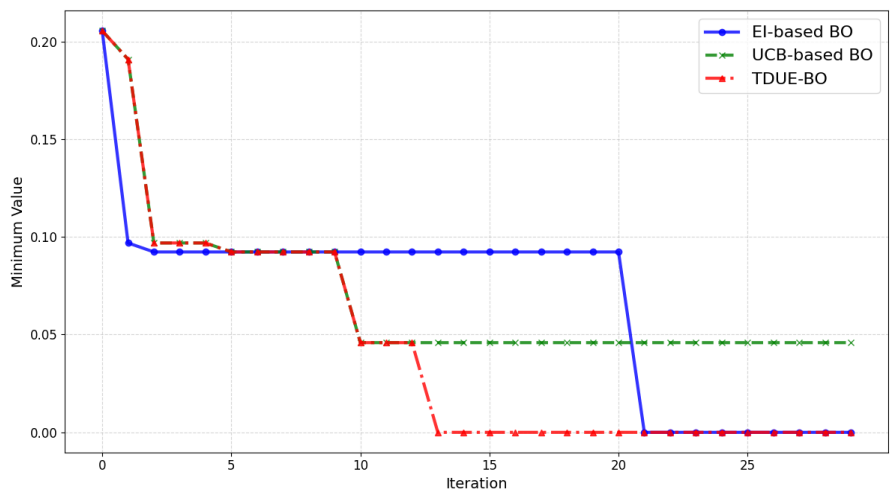

Figure 13: Iterations to reach the minimum for the three competing approaches for the AutoAM dataset

## 5. Conclusion

In this work, we have proposed a novel approach of performing the sequential experiments using the BO active learning framework. We have augmented the performance of the classical BO, which uses EI or UCB as the means to guide the iterative process of choosing subsequent samples, by a threshold guided method based on the uncertainty quantified by the GP model. By taking advantage of the strengths of UCB and EI, our proposed method overcomes the weaknesses of the UCB and EI-based BO. In TDUE-BO, the use of UCB at the outset capitalizes on its exploratory strengths, ensuring a comprehensive initial sweep of the MDS. This is crucial in material science applications where the objective functions often exhibit multi-modal and intricate behaviors. As our GP model accrues data and refines its understanding, reflected by a decrease in the average standard deviation of predictions, we seamlessly transition to the EI acquisition function. This shift marks a transition to a more exploitation-focused phase, concentrating our search around the most promising areas identified during the initial exploration. We have employed our proposed methodology in three different datasets from the field of material science and the results have reflected the advantages of this adaptive approach. We observed a better realization of the relationship between the process parameters and the target output using the TDUE-BO approach in all three datasets compared to conventional BO methods that rely on a single type of acquisition function throughout. The threshold-driven switch from UCB to EI ensures that our method not only approximates the MDS efficiently but also converges to the optimal region with lower number of sequential updates, adapting its focus from exploration to exploitation. This adaptability is particularly beneficial in material science, where the diversity of datasets and the complexity of their underlying phenomena demand a flexible optimization approach.

Future directions of this work could include trying our proposed framework with high-dimensional datasets as the performance of BO continues to deteriorate when the dimension of the input features increases. Besides, since BO is often criticized for being myopic in nature, reinforcement learning (RL) could be incorporated to achieve the multi-step lookahead capability in BO. RL could also be employed to choose a different acquisition function for each sequential update from a large set of available acquisition functions to get a much better convergence and approximation.

## Acknowledgements

We extend our sincere gratitude to the Department of Industrial and Management Systems Engineering (IMSE) for their unwavering support throughout the duration of this research. Additionally, we are deeply grateful to our fellow professors and researchers specializing in the field of additive manufacturing. Their collaborative spirit, shared expertise, and insightful feedback have significantly enriched this study.

## Acknowledgements

We extend our sincere gratitude to the Department of Industrial and Management Systems Engineering (IMSE) for their unwavering support throughout the duration of this research. Additionally, we are deeply grateful to our fellow professors and researchers specializing in the field of additive manufacturing. Their collaborative spirit, shared expertise, and insightful feedback have significantly enriched this study.

## References

[1] E. O. Pyzer-Knapp et al., "Accelerating materials discovery using artificial intelligence, high performance computing and robotics," *npj Comput. Mater.*, vol. 8, no. 1, p. 84, 2022, doi: 10.1038/s41524-022-00765-z.

[2] Y. Juan, Y. Dai, Y. Yang, and J. Zhang, "Accelerating materials discovery using machine learning," *J. Mater. Sci. Technol.*, vol. 79, pp. 178–190, 2021, doi: https://doi.org/10.1016/j.jmst.2020.12.010.

[3] Q. Luo et al., "Thermodynamics and kinetics of phase transformation in rare earth–magnesium alloys: A critical review," *J. Mater. Sci. Technol.*, vol. 44, pp. 171–190, 2020, doi: https://doi.org/10.1016/j.jmst.2020.01.022.

[4] X. Wu, F. Kang, W. Duan, and J. Li, "Density functional theory calculations: A powerful tool to simulate and design high-performance energy storage and conversion materials," *Prog. Nat. Sci. Mater. Int.*, vol. 29, no. 3, pp. 247–255, 2019, doi: https://doi.org/10.1016/j.pnsc.2019.04.003.

[5] P. Nikolaev et al., "Autonomy in materials research: a case study in carbon nanotube growth," *npj Comput. Mater.*, vol. 2, no. 1, p. 16031, 2016, doi: 10.1038/npjcompumats.2016.31.

[6] A. G. Kusne et al., "On-the-fly closed-loop materials discovery via Bayesian active learning," *Nat. Commun.*, vol. 11, no. 1, p. 5966, 2020, doi: 10.1038/s41467-020-19597-w.

[7] H. C. Herbol, M. Poloczek, and P. Clancy, "Cost-effective materials discovery: Bayesian optimization across multiple information sources," *Mater. Horizons*, vol. 7, no. 8, pp. 2113–2123, 2020, doi: 10.1039/D0MH00062K.

[8] A. S. Raihan and I. Ahmed, "Guiding the Sequential Experiments in Autonomous Experimentation Platforms through EI-based Bayesian Optimization and Bayesian Model Averaging," 2023, doi: https://doi.org/10.48550/arXiv.2302.13360.

[9] S. Diwale, M. K. Eisner, C. Carpenter, W. Sun, G. C. Rutledge, and R. D. Braatz, "Bayesian optimization for material discovery processes with noise," *Mol. Syst. Des. Eng.*, vol. 7, no. 6, pp. 622–636, 2022.

[10] J. Zhang, X. Li, D. Xu, and R. Yang, "Recent progress in the simulation of microstructure evolution in titanium alloys," *Prog. Nat. Sci. Mater. Int.*, vol. 29, no. 3, pp. 295–304, 2019, doi: https://doi.org/10.1016/j.pnsc.2019.05.006.

[11] W. Zhu, Y. Xu, J. Ni, G. Hu, X. Wang, and W. Zhang, "SEHC: A high-throughput materials computing framework with automatic self-evaluation filtering," *Mater. Sci. Eng. B*, vol. 252, p. 114474, 2020, doi: https://doi.org/10.1016/j.mseb.2019.114474.

[12] J. H. Montoya, K. T. Winther, R. A. Flores, T. Bligaard, J. S. Hummelshøj, and M. Aykol, "Autonomous intelligent agents for accelerated materials discovery," *Chem. Sci.*, vol. 11, no. 32, pp. 8517–8532, 2020.

[13] S. T. S. Bukkapatnam, "Autonomous materials discovery and manufacturing (AMDM): A review and perspectives," *IISE Trans.*, vol. 55, no. 1, pp. 75–93, Jan. 2023, doi: 10.1080/24725854.2022.2089785.

[14] B. Lei et al., "Bayesian optimization with adaptive surrogate models for automated experimental design," *npj Comput. Mater.*, vol. 7, no. 1, p. 194, 2021, doi: 10.1038/s41524-021-00662-x.

[15] A. Talapatra, S. Boluki, T. Duong, X. Qian, E. Dougherty, and R. Arróyave, "Autonomous efficient experiment design for materials discovery with Bayesian model averaging," *Phys. Rev. Mater.*, vol. 2, no. 11, 2018, doi: 10.1103/PhysRevMaterials.2.113803.

[16] D. R. Jones, M. Schonlau, and W. J. Welch, "Efficient Global Optimization of Expensive Black-Box Functions," *J. Glob. Optim.*, vol. 13, no. 4, pp. 455–492, 1998, doi: 10.1023/A:1008306431147.

[17] P. I. Frazier and J. Wang, "Bayesian optimization for materials design," *Springer Ser. Mater. Sci.*, vol. 225, pp. 45–75, 2015, doi: 10.1007/978-3-319-23871-5_3.

[18] Q. Liang et al., "Benchmarking the performance of Bayesian optimization across multiple experimental materials science domains," *npj Comput. Mater.*, vol. 7, no. 1, p. 188, 2021, doi: 10.1038/s41524-021-00656-9.

[19] B. Shahriari, K. Swersky, Z. Wang, R. P. Adams, and N. de Freitas, "Taking the Human Out of the Loop: A Review of Bayesian Optimization," *Proc. IEEE*, vol. 104, no. 1, pp. 148–175, 2016, doi: 10.1109/JPROC.2015.2494218.

[20] P. I. Frazier, "A Tutorial on Bayesian Optimization," *arXiv Prepr. arXiv1807.02811*, 2018, doi:

https://doi.org/10.48550/arXiv.1807.02811.
[21] E. Brochu, V. M. Cora, and N. de Freitas, "A Tutorial on Bayesian Optimization of Expensive Cost Functions, with Application to Active User Modeling and Hierarchical Reinforcement Learning," 2010, [Online]. Available: http://arxiv.org/abs/1012.2599.
[22] M. M. Flores-Leonar *et al.*, "Materials Acceleration Platforms: On the way to autonomous experimentation," *Curr. Opin. Green Sustain. Chem.*, vol. 25, p. 100370, 2020, doi: https://doi.org/10.1016/j.cogsc.2020.100370.
[23] E. O. Pyzer-Knapp, "Bayesian optimization for accelerated drug discovery," *IBM J. Res. Dev.*, vol. 62, no. 6, pp. 2:1-2:7, 2018, doi: 10.1147/JRD.2018.2881731.
[24] H. Bellamy, A. A. Rehim, O. I. Orhobor, and R. King, "Batched Bayesian Optimization for Drug Design in Noisy Environments," *J. Chem. Inf. Model.*, vol. 62, no. 17, pp. 3970–3981, Sep. 2022, doi: 10.1021/acs.jcim.2c00602.
[25] L. Colliandre and C. Muller, "Bayesian Optimization in Drug Discovery BT - High Performance Computing for Drug Discovery and Biomedicine," A. Heifetz, Ed. New York, NY: Springer US, 2024, pp. 101–136.
[26] F. Berkenkamp, A. Krause, and A. P. Schoellig, "Bayesian optimization with safety constraints: safe and automatic parameter tuning in robotics," *Mach. Learn.*, vol. 112, no. 10, pp. 3713–3747, 2023, doi: 10.1007/s10994-021-06019-1.
[27] K. Junge, J. Hughes, T. G. Thuruthel, and F. Iida, "Improving Robotic Cooking Using Batch Bayesian Optimization," *IEEE Robot. Autom. Lett.*, vol. 5, no. 2, pp. 760–765, 2020, doi: 10.1109/LRA.2020.2965418.
[28] R. Lam, M. Poloczek, P. Frazier, and K. E. Willcox, "Advances in Bayesian Optimization with Applications in Aerospace Engineering," in *2018 AIAA Non-Deterministic Approaches Conference*, American Institute of Aeronautics and Astronautics, 2018.
[29] A. Hebbal, L. Brevault, M. Balesdent, E.-G. Talbi, and N. Melab, "Bayesian optimization using deep Gaussian processes with applications to aerospace system design," *Optim. Eng.*, vol. 22, no. 1, pp. 321–361, 2021, doi: 10.1007/s11081-020-09517-8.
[30] A. Pal, L. Zhu, Y. Wang, and G. G. Zhu, "Multi-Objective Stochastic Bayesian Optimization for Iterative Engine Calibration," in *2020 American Control Conference (ACC)*, 2020, pp. 4893–4898, doi: 10.23919/ACC45564.2020.9147983.
[31] L. Zhu, Y. Wang, A. Pal, and G. G. Zhu, "Adaptive Design of Experiments for Automotive Engine Applications Using Concurrent Bayesian Optimization," *ASME Lett. Dyn. Syst. Control*, vol. 2, no. 3, Apr. 2022, doi: 10.1115/1.4054222.
[32] K. Wang and A. W. Dowling, "Bayesian optimization for chemical products and functional materials," *Curr. Opin. Chem. Eng.*, vol. 36, p. 100728, 2022, doi: https://doi.org/10.1016/j.coche.2021.100728.
[33] S. Park, J. Na, M. Kim, and J. M. Lee, "Multi-objective Bayesian optimization of chemical reactor design using computational fluid dynamics," *Comput. Chem. Eng.*, vol. 119, pp. 25–37, 2018, doi: https://doi.org/10.1016/j.compchemeng.2018.08.005.
[34] R. R. Griffiths and J. M. Hernández-Lobato, "Constrained Bayesian optimization for automatic chemical design using variational autoencoders," *Chem. Sci.*, vol. 11, no. 2, pp. 577–586, 2020.
[35] H.-E. Byun, B. Kim, and J. H. Lee, "Multi-step lookahead Bayesian optimization with active learning using reinforcement learning and its application to data-driven batch-to-batch optimization," *Comput. Chem. Eng.*, vol. 167, p. 107987, 2022, doi: https://doi.org/10.1016/j.compchemeng.2022.107987.
[36] I. Ahmed, S. Bukkapatnam, B. Botcha, and Y. Ding, "Towards Futuristic Autonomous Experimentation--A Surprise-Reacting Sequential Experiment Policy," pp. 1–25, 2021, [Online]. Available: https://arxiv.org/abs/2112.00600v1.
[37] A. D. Bull, "Convergence rates of efficient global optimization algorithms," *J. Mach. Learn. Res.*, vol. 12, pp. 2879–2904, 2011.
[38] Z. Chen, S. Mak, and C. F. J. Wu, "A Hierarchical Expected Improvement Method for Bayesian Optimization," *J. Am. Stat. Assoc.*, pp. 1–14, 2023, doi: 10.1080/01621459.2023.2210803.
[39] D. Bash *et al.*, "Multi-Fidelity High-Throughput Optimization of Electrical Conductivity in P3HT-CNT Composites," *Adv. Funct. Mater.*, vol. 31, no. 36, p. 2102606, Sep. 2021, doi: https://doi.org/10.1002/adfm.202102606.
[40] S. Sun *et al.*, "A data fusion approach to optimize compositional stability of halide perovskites," *Matter*, vol. 4, no. 4, pp. 1305–1322, 2021, doi: https://doi.org/10.1016/j.matt.2021.01.008.
[41] J. R. Deneault *et al.*, "Toward autonomous additive manufacturing: Bayesian optimization on a 3D printer," *MRS Bull.*, vol. 46, no. 7, pp. 566–575, 2021, doi: 10.1557/s43577-021-00051-1.
[42] H. Abdi and L. J. Williams, "Principal component analysis," *WIREs Comput. Stat.*, vol. 2, no. 4, pp. 433–459, Jul. 2010, doi: https://doi.org/10.1002/wics.101.

[43] C. E. Rasmussen and C. K. Williams, *Gaussian processes for machine learning*. Cambridge, MA: MIT press, 2006.

[44] S. Jin, A. Iquebal, S. Bukkapatnam, A. Gaynor, and Y. Ding, "A Gaussian Process Model-Guided Surface Polishing Process in Additive Manufacturing," *J. Manuf. Sci. Eng.*, vol. 142, no. 1, Nov. 2019, doi: 10.1115/1.4045334.

[45] K. Kawaguchi, L. P. Kaelbling, and T. Lozano-Pérez, "Bayesian Optimization with Exponential Convergence," in *Advances in Neural Information Processing Systems*, 2015, vol. 28, [Online]. Available: https://proceedings.neurips.cc/paper_files/paper/2015/file/0ebcc77dc72360d0eb8e9504c78d38bd-Paper.pdf.

[46] G. De Ath, R. M. Everson, A. A. M. Rahat, and J. E. Fieldsend, "Greed Is Good: Exploration and Exploitation Trade-Offs in Bayesian Optimisation," *ACM Trans. Evol. Learn. Optim.*, vol. 1, no. 1, Apr. 2021, doi: 10.1145/3425501.

[47] X. Zhang, S. Das, and K. Kreutz-Delgado, "Tuning Confidence Bound for Stochastic Bandits with Bandit Distance," *arXiv Prepr. arXiv2110.02690*, 2021, doi: https://doi.org/10.48550/arXiv.2110.02690.